\documentclass[acmtog]{acmart}
\acmSubmissionID{561}

\usepackage{booktabs} 
\usepackage{svg}
\usepackage{hyperref}
\usepackage{graphicx}
\usepackage[ruled]{algorithm2e} 
\usepackage{algpseudocode}
\SetAlFnt{\small}
\SetAlCapFnt{\small}
\SetAlCapNameFnt{\small}
\SetAlCapHSkip{0pt}
\acmJournal{TOG}

\copyrightyear{2024}
\acmYear{2024}
\setcopyright{rightsretained}
\acmConference[SA Conference Papers '24]{SIGGRAPH Asia 2024 Conference Papers}{December 3--6, 2024}{Tokyo, Japan}
\acmBooktitle{SIGGRAPH Asia 2024 Conference Papers (SA Conference Papers '24), December 3--6, 2024, Tokyo, Japan}\acmDOI{10.1145/3680528.3687623}
\acmISBN{979-8-4007-1131-2/24/12}

\begin{document}
\title{Fashion-VDM: Video Diffusion Model for Virtual Try-On}
\newcommand{\FashionVDM}{Fashion-VDM\xspace}

\author{Johanna Karras}
\orcid{0009-0003-9999-5510}
\affiliation{%
  \institution{Google Research, University of Washington}
  \country{USA}
}
\email{jskarras@cs.washington.edu}

\author{Yingwei Li}
\orcid{0009-0002-2968-0735}
\affiliation{%
  \institution{Google Research}
  \country{USA}
}
\email{yingweili@google.com}

\author{Nan Liu}
\orcid{0000-0002-1380-5428}
\affiliation{%
  \institution{Google Research}
  \country{USA}
}
\email{naanliu@google.com}

\author{Luyang Zhu}
\orcid{0009-0003-8543-9177}
\affiliation{%
  \institution{Google Research, University of Washington}
  \country{USA}
}
\email{lyzhu@cs.washington.edu}

\author{Innfarn Yoo}
\orcid{0000-0003-4616-4644}
\affiliation{%
  \institution{Google Research}
  \country{USA}
}
\email{innfarn@google.com}

\author{Andreas Lugmayr}
\orcid{0000-0001-5154-6116}
\affiliation{%
  \institution{Google Research}
  \country{USA}
}
\email{alugmayr@google.com}

\author{Chris Lee}
\orcid{0009-0005-5703-9605}
\affiliation{%
  \institution{Google Research}
  \country{USA}
}
\email{chrisalee@google.com}

\author{Ira Kemelmacher-Shlizerman}
\orcid{0009-0003-9498-584X}
\affiliation{%
  \institution{Google Research, University of Washington}
  \country{USA}
}
\email{kemelmi@google.com}

\renewcommand\shortauthors{Karras, J. et al}

\begin{abstract}
We present Fashion-VDM, a video diffusion model (VDM) for generating virtual try-on videos. Given an input garment image and person video, our method aims to generate a high-quality try-on video of the person wearing the given garment, while preserving the person's identity and motion. Image-based virtual try-on has shown impressive results; however, existing video virtual try-on (VVT) methods are still lacking garment details and temporal consistency. To address these issues, we propose a diffusion-based architecture for video virtual try-on, split classifier-free guidance for increased control over the conditioning inputs, and a progressive temporal training strategy for single-pass 64-frame, 512px video generation. We also demonstrate the effectiveness of joint image-video training for video try-on, especially when video data is limited. Our qualitative and quantitative experiments show that our approach sets the new state-of-the-art for video virtual try-on. For additional results, visit our project page: \href{https://johannakarras.github.io/Fashion-VDM}{https://johannakarras.github.io/Fashion-VDM}.
\end{abstract}

%
%
%
%


\begin{teaserfigure}
  \centering
  \includegraphics[width=0.9\textwidth]{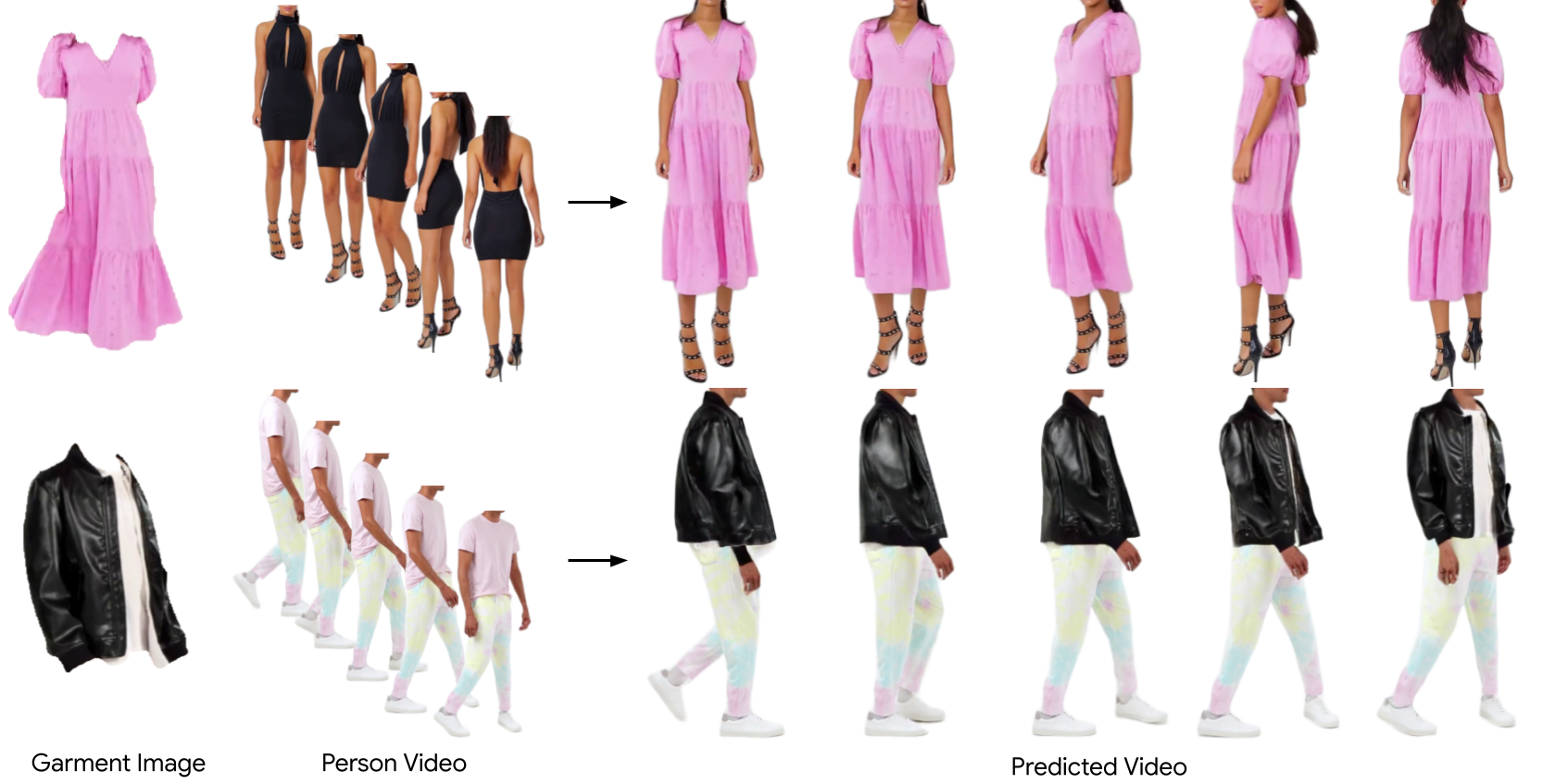}
  \caption{Fashion-VDM. Given an input garment image and a person video, Fashion-VDM generates a video of the person virtually trying on the given garment, while preserving their original identity and motion.}
  \label{fig:teaser}
\end{teaserfigure}


\maketitle

\section{Introduction}
    With the popularity of online clothing shopping and social media marketing, there is a strong demand for virtual try-on methods. Given a garment image and a person image, virtual try-on aims to show how the person would look wearing the given garment. In this paper, we explore \textit{video} virtual try-on, where the input is a garment image and person video. The benefit of a video virtual try-on (VVT) experience is that it would depict how a garment looks at different angles and how it drapes and flows in motion. 

    VVT is a challenging task, as it requires synthesizing realistic try-on frames from different viewpoints, while generating realistic fabric dynamics (e.g. folds and wrinkles) and maintaining temporal consistency between frames. Additional difficulty arises if the person and garment poses vary significantly, as this creates occluded garment and person regions that need to be hallucinated. Another challenge is the scarcity of try-on video data. Perfect ground truth data (i.e. two videos of different people wearing the same garment and moving in the exact same way) is difficult and expensive to acquire. In general, available human video data, such as UBC Fashion ~\cite{ubc-dataset}, are much more scarce and less diverse than image data, such as LAION 5B~\cite{laion5b}.
    
    Past approaches to virtual try-on typically leverage dense flow fields to explicitly warp the source garment pixels onto the target person frames \cite{attn-vvt, mv-ton, clothformer, dressing-in-the-wild, fw-gan}. However, these flow-based approaches can introduce artifacts due to occlusions in the source frame, large pose deformations, and inaccurate flow estimates. Moreover, these methods are incapable of producing realistic and fine-grained fabric dynamics, such as wrinkling, folding, and flowing, as these details are not captured by appearance flows. A recent breakthrough in image-based virtual try-on uses a diffusion model~\cite{tryondiffusion}, which implicitly warps the input garment under large pose gaps and heavy occlusion using spatial cross-attention. However, directly applying \cite{tryondiffusion} or other image-based try-on methods for VVT in a frame-by-frame manner creates severe flickering artifacts and temporal inconsistencies. 
    
    
    Diffusion models~\cite{diffusion-model-1, diffusion-model-2, diffusion-model-3, ddpm, ddim} have shown promising results on various video synthesis tasks, such as text-to-video generation~\cite{vdm} and image-to-video generation~\cite{dreampose, animate-anyone, animate-diff}. However, a key challenge is generating longer videos, while maintaining temporal consistency and adhering to computational and memory constraints. Previous works use cascaded approaches~\cite{imagen-video}, sliding windows inference~\cite{vdm, magic-animate}, past-frame conditioning~\cite{flexible-dm-long-videos,soundini,vidm}, and transitions or interpolation~\cite{seine,genlvideo}. Yet, even with such schemes, longer videos are temporally inconsistent, contain artifacts, and lack realistic textures and details. We argue that, similar to context modeling for LLM's~\cite{context-modeling-llm}, short-video generation models can be naturally extended for long-video generation by a temporally progressive finetuning scheme, without introducing additional inference passes or multiple networks.
    
    A potential option for diffusion-based VVT is to apply an animation model to a single try-on image generated by an image try-on model. However, as this is not an end-to-end trained system, any image try-on errors will accumulate throughout the video. 
    We argue that a single VVT model would overcome this issue by 1) injecting explicit person and garment conditioning information into the model and 2) having an end-to-end training objective.

    We present \textit{Fashion-VDM}, the first VVT method to synthesize temporally consistent, high-quality try-on videos, even on diverse poses and difficult garments. Fashion-VDM is a single-network, diffusion-based approach. To maintain temporal smoothness, we inflate the M\&M VTO~\cite{mixmatch} architecture with 3D-convolution and temporal attention blocks. We maintain temporal consistency in videos up to 64-frames long with a single network by training in a temporally progressive manner. To address input person and garment fidelity, we introduce split classifier-free guidance (split-CFG) that enables increased control over each input signal. In our experiments, we also show that split-CFG increases realism, temporal consistency, and garment fidelity, compared to ordinary or dual CFG. Additionally, we increase garment fidelity and realism by training jointly with image and video data. Our results show that Fashion-VDM surpasses benchmark methods by a large margin and synthesizes state-of-the-art try-on videos.


    \begin{figure*}
  \centering
  \includegraphics[width=\linewidth]{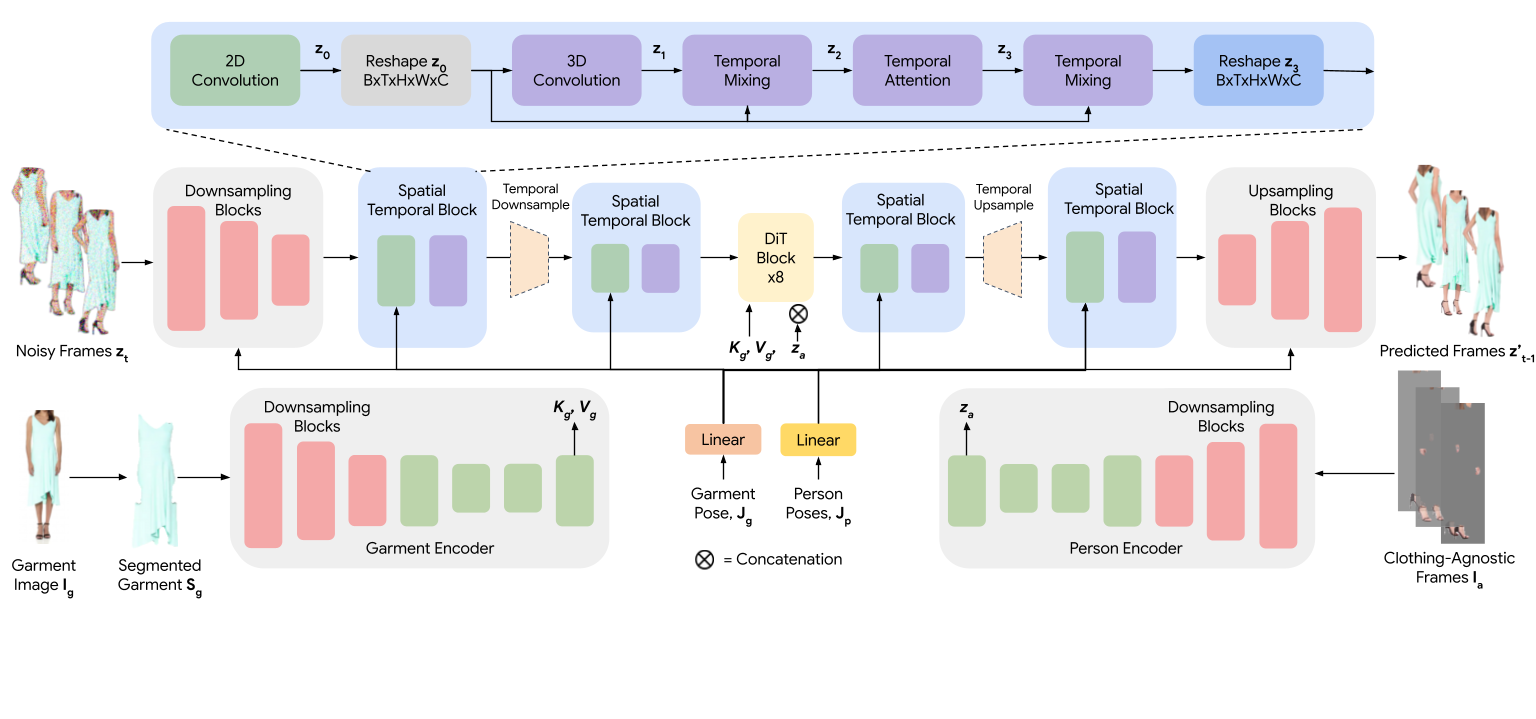}
  \vspace{-50pt}
  \caption{Fashion-VDM Architecture. Given a noisy video $z_t$ at diffusion timestep $t$, a forward pass of Fashion-VDM computes a single denoising step to get the denoised video $z'_{t-1}$. Noisy video $z_t$ is preprocessed into person poses $J_p$ and clothing-agnostic frames $I_a$, while the garment image $I_g$ is preprocessed into the garment segmentation $S_g$ and garment poses $J_g$ (Section ~\ref{ssec:input-preprocessing}). The architecture follows~\cite{mixmatch}, except the main UNet contains  3D-Conv and temporal attention blocks to maintain temporal consistency. Additionally, we inject temporal down/upsampling blocks during 64-frame temporal training. Noisy video $z_t$ is encoded by the main UNet and the conditioning signals, $S_g$ and $I_a$, are encoded by separate UNet encoders. In the 8 DiT blocks at the lowest resolution of the UNet, the garment conditioning features are cross-attended with the noisy video features and the spatially-aligned clothing-agnostic features $z_a$ and noisy video features are directly concatenated. $J_g$ and $J_p$ are encoded by single linear layers, then concatenated to the noisy features in all UNet 2D spatial layers. }
  \label{fig:architecture}
\end{figure*}

\section{Related Works}

\subsection{Video Diffusion Models}
    Many early video diffusion models~\cite{vdm} (VDMs) adapt text-to-image diffusion models to generate batches of consecutive video frames, often employing temporal blocks within the denoising UNet architecture to learn temporal consistency~\cite{vdm, imagen-video}. Latent VDM's~\cite{vidm, align-your-latents, reuse-and-diffuse, animate-diff, dreampose, long-videos, stable-video-diffusion, lavie} reduce the computational complexity of standard VDM's by performing diffusion in the latent space.

    To achieve longer videos and increased spatial resolution, \cite{imagen-video} proposes a cascade of temporal and spatial upsampling UNets. Other methods employ similar schemes of cascaded models for long video generation~\cite{lavie}. However, cascaded strategies require multiple networks and inference runs. Another strategy is to synthesize sparse keyframes, then use frame interpolation~\cite{vidm}, past-frame conditioning~\cite{long-videos}, temporally overlapping frames~\cite{magic-animate}, and predicting transitions between frames~\cite{seine, genlvideo} to achieve longer, temporal-consistent videos. 
    Unlike past long-video VDM's, Fashion-VDM is a unified (non-cascaded) diffusion model that generates a long video up to 64 frames long in a single inference run, thereby reducing memory requirements and inference time.

\subsection{Image and Pose Guidance}

    Many VDM's are text-conditioned~\cite{vidm, imagen-video, align-your-latents, stable-video-diffusion} and there is increasing interest in image-conditioned VDM's~\cite{dreampose, animate-anyone, animate-diff}. 
    To maintain the exact details of input images, some methods require inference-time finetuning~\cite{dreampose, stable-video-diffusion, animate-diff}.
    In contrast, Fashion-VDM requires no additional finetuning during test time to maintain high-quality details of the input person and garment.

    Some recent diffusion-based animation methods are both image- and pose-conditioned~\cite{dreampose, emu-video, animate-anyone, animate-diff, magic-animate}.
    DreamPose uses a pre-trained (latent) Stable Diffusion model without temporal layers to generate videos in a frame-by-frame manner~\cite{dreampose}. More recently, Animate Anyone~\cite{animate-anyone} encodes the image using ReferenceNet and their diffusion model incorporates spatial, cross, and temporal attention layers to maintain consistency and preserve details, while MagicAnimate~\cite{magic-animate} introduces an appearance encoder to maintain the fidelity across the frames and generates a long video using temporally overlapping segments. 
    In contrast, Fashion-VDM is a non-latent, temporally-aware video diffusion model, capable of synthesizing up to 64 consecutive frames in a single inference pass.

\subsection{Virtual Try-On}
    
    Traditional image virtual try-on approaches first warp the target garment onto the input person, then refine the resulting image \cite{viton, viton-hd, hr-viton, single-stage-viton, global-appearance-flow, controllable-viton, neural-texture-viton, photo-realistic-viton, vtnfp, pise, street-tryon}. Similarly, for video virtual try-on (VVT), past methods often rely on multiple networks to predict intermediate values, such as optical flow, background masks, and occlusion masks, to warp the target garment to the person in each frame of the video~\cite{attn-vvt, clothformer, dressing-in-the-wild, fw-gan, mv-ton}. However, inaccuracies in these intermediate values lead to artifacts and misalignment. Some image try-on approaches incorporate optical flow estimation to alleviate this misalignment~\cite{single-stage-viton, hr-viton, tryon-gan, clothflow}. For VVT, MV-TON~\cite{mv-ton} proposes a memory refinement module to correct inaccurate details in the generated frames by encoding past frames into latent space, then using this as external memory to generate new frames. ClothFormer~\cite{clothformer} estimates an occlusion mask to correct for flow inaccuracies. Current state-of-the-art VVT methods achieve improved results by utilizing attention modules in the warping and fusing phases~\cite{clothformer, attn-vvt}.

    In contrast to earlier flow-based methods, TryOnDiffusion~\cite{tryondiffusion} leverages a diffusion-based method conditioned with pose and garment for image virtual try-on. 
    WarpDiffusion~\cite{warp-diffusion} tries to reduce the computational cost and data requirements by bridging warping and diffusion-based virtual try-on methods. StableVITON~\cite{stable-viton} avoids warping by finetuning pre-trained latent diffusion~\cite{ldm} encoders for input person and garment conditioning via cross-attention blocks. Mix-and-match (M\&M) VTO~\cite{tryondiffusion} extends single tryon task for mixmatch tryon application with a novel person embedding finetuning strategy.


\subsection{Image and Video Training}
    Video datasets are often smaller and less diverse, compared to image datasets, as images are more abundant online. To alleviate this problem, \cite{vdm, imagen-video, magic-animate} propose jointly leveraging image and video data for training. VDM~\cite{vdm} and Imagen Video~\cite{imagen-video} implement joint training by applying a temporal mask to image batches. MagicAnimate~\cite{magic-animate} applies joint training during the pretraining stage of their appearance encoder and pose ControlNet. We improve upon existing joint training schemes (see Section~\ref{ssec:joint-training}), ultimately demonstrating the benefit of joint image and video training for video try-on.

    \begin{figure}
  \centering
  \includegraphics[width=\linewidth]{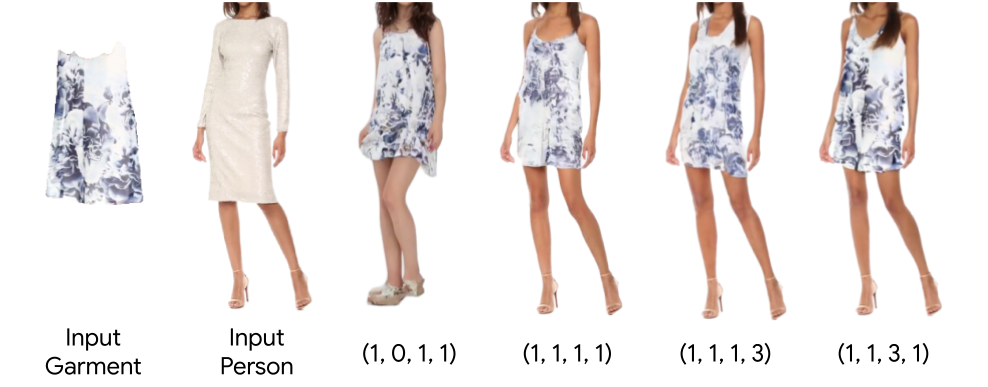}
  \vspace{-20pt}
  \caption{Split-CFG Ablation. We compare different split-cfg weights, where $(w_{\emptyset}, w_\text{p}, w_\text{g}, w_\text{full})$ correspond to the unconditional guidance, person-only guidance, person and cloth guidance, and full guidance terms, respectively. }
  \vspace{-10pt}
  \label{fig:split-cfg-1}
\end{figure}
    
\section{Method}
    We propose Fashion-VDM, a unified video diffusion model for synthesizing state-of-the-art virtual try-on (VTO) videos up to 64 frames long at $512$px resolution. Our method introduces an end-to-end diffusion-based VVT architecture based on \cite{mixmatch} (Section \ref{ssec:architecture}), split classifier-free guidance (split-CFG) for increased garment fidelity (Section \ref{ssec:split-cfg}),  progressive temporal training for long-video generation (Section \ref{ssec:progressive-training}), and joint image-video training for improved garment fidelity (Section \ref{ssec:joint-training}).

    \subsection{Problem Formulation} \label{ssec:problem-formulation}
    In video virtual try-on, the input is a video $\{I^0_{p}, I^1_{p}, ..., I^{N-1}_{p}\}$  of a person $p$ consisting of $N$ frames and a single garment image $I_g$ of another person wearing garment $g$. The goal is to synthesize a video $\{I^0_{tr}, I^1_{tr}, ..., I^{N-1}_{tr}\}$, where $I^i_{tr}$ denotes the $i$-th \textbf{tr}y-on video frame that preserves the identity and motion of the person $p$ wearing the garment $g$. 
    
    \subsection{Preliminary: M\&M VTO} \label{ssec:preliminaries}
    
    Our VTO-UDiT network architecture is inspired by~\cite{mixmatch}, a state-of-the-art multi-garment \textit{image} try-on diffusion model that also enables text-based control of garment layout. VTO-UDiT is represented by
    \begin{equation}
        \hat{x_0} = x_{\theta}(z_t, t, c_{tr})
    \end{equation}
    where $\hat{x_0}$ is the predicted try-on image by the network $x_{\theta}$, parameterized by $\theta$, at diffusion timestep $t$, $z_t$ is the noisy image, and $c_{tr}$ is the conditioning inputs. VTO-UDiT is parameterized in v-space, following~\cite{v_prediction}. Each conditioning input is encoded separately by fully convolutional encoders and processed at the lowest resolution of the main UNet via DiT blocks~\cite{DiT}, where conditioning features are processed with self-attention or cross-attention modules. However, while it shows impressive results for image try-on,  VTO-UDiT cannot reason about temporal consistency when applied to video inputs.
    
    \subsection{Input Preprocessing} \label{ssec:input-preprocessing} 
    From the input video frames, we compute the clothing-agnostic frames $I_a = \{I^0_{a}, I^1_{a}, ..., I^{N-1}_{a}\}$, person poses $J_p = \{J^0_{p}, J^1_{p}, ..., J^{N-1}_{p}\}$, and person masks $\{M^0_p, M^1_p, ..., M^{N-1}_p\}$. The clothing-agnostic frames mask out the entire bounding box area of the person in the frame, except for the visible body regions (head, hands, legs, and shoes), following TryOnDiffusion~\cite{tryondiffusion}. 
    Optionally, the clothing-agnostic frames can keep the original bottoms, if doing top try-on only.
    From the input garment image $I_g$, we extract the garment segmentation image $S_g$, garment pose $J_g$, and garment mask $M_g$. The garment pose refers to the pose keypoints of the person wearing the garment before segmentation. We channel-wise concatenate $M^i_p$ to $I^i_a$ and $M_g$ to $I_g$. Poses, masks, and segmentations are computed using an in-house equivalent of Graphonomy~\cite{graphonomy}. 
    Both person and garment pose keypoints are preprocessed to be spatially aligned with the person frames and garment image, respectively. 
        
    \subsection{Architecture} \label{ssec:architecture}
        Our overall architecture is depicted in Figure \ref{fig:architecture}. 
        We adapt the VTO-UDiT architecture \cite{tryondiffusion} by inflating the two lowest-resolution downsampling and upsampling blocks with temporal attention and 3D-Conv blocks, as shown in Figure~\ref{fig:architecture}. To be specific, after the 2D-Conv layers, we add a 3D-Conv block, a temporal attention block, and a temporal mixing block to linearly combine spatial and temporal features, as proposed in \cite{align-your-latents}. In the temporal mixing blocks, processed features after the spatial attention layer $z_s$ are linearly combined with processed features after the temporal attention layer $z_t$ via learned weighting parameter $\alpha$:
        
        \begin{equation}
           z'_t = \alpha \cdot z_s + (1 - \alpha) \cdot z_t
        \end{equation}

        During 64-frame training (see Section~\ref{ssec:progressive-training}), we further inflate the model with temporal downsampling and upsampling blocks with factor 2, to reduce the memory footprint of the model. These blocks are added before and after the lowest-resolution spatial blocks, respectively.

        The person and garment poses are encoded and used to condition all 2D spatial layers in the UNet. The 8 Diffusion Transformer (DiT) blocks~\cite{DiT} between the UNet encoder and decoder condition our model on the segmented garment and clothing-agnostic image features, as proposed by \cite{mixmatch}. In each block, the garment images are cross-attended with the noisy target features, while the agnostic input images are concatenated to the noisy target features.
    
    \subsection{Split Classifier-Free Guidance}\label{ssec:split-cfg}
        Standard classifier-free guidance (CFG)~\cite{classifier-free-guidance} is a sampling technique that pushes the distribution of inference results towards the input conditioning signal(s); however, it does not allow for disentangled guidance towards separate conditioning signals. Instruct-Pix2Pix~\cite{instruct-pix2pix} introduces dual-CFG, which separates the CFG weights for text and image conditioning signals, drawing inspiration from Composable Diffusion~\cite{liu2022compositional}.

        We introduce \textit{split}-CFG, a generalization of dual-CFG which allows independent control over multiple conditioning signals. See Algorithm~\ref{split-cfg}. The inputs to \texttt{Split-CFG} are the trained denoising UNet $\epsilon_{\theta}$, the list of all conditioning signal sets $C$, and the respective conditioning weights $W$. For each subset of conditioning signals $c_i \in C$, containing one or more conditional inputs, the algorithm computes the conditional result $\epsilon_i$ given $c_i$. Then, the weighted difference between the conditional result $\epsilon_i$ from the past conditional result $\epsilon_{i-1}$ is added to the prediction. In this way, the prediction is pushed in the direction of $c_i$.
    
        \begin{algorithm}[hbt!]
        \caption{Split Classifier-Free Guidance}\label{split-cfg}
        \SetAlgoLined
        \SetKwProg{myalg}{Split-CFG }{$(\epsilon_{\theta}, C, W)$}{}
        \myalg{}{
        $c \gets \emptyset $\Comment{ current conditioning signals}\;
        $\hat{\epsilon}_{\theta}(z_t, C) \gets  w_0 \epsilon_{\theta}(z_t, \emptyset)$\Comment{initialize prediction}\;
        $\hat{\epsilon}_{0} \gets \hat{\epsilon}_{\theta}(z_t, C)$\Comment{store past prediction}\;
          \For{$c_i$ in C}{
            $c \gets c \cup \{c_i\}$ \Comment{update $c$}\;
            $\hat{\epsilon}_i  \gets \epsilon_{\theta}(z_t, c)$\Comment{store new prediction}\;
            $\hat{\epsilon}_{\theta}(z_t, C) \gets \hat{\epsilon}_{\theta}(z_t, C) + w_i (\hat{\epsilon}_i - \hat{\epsilon}_{i-1})$ \;
            $\hat{\epsilon}_{i-1} \gets \hat{\epsilon}_{i}$ \Comment{update $\hat{\epsilon}_{i-1}$}
         }
        \textbf{return $\hat{\epsilon}_{\theta}(z_t, C)$}
            
        }
        \end{algorithm}
    
        Split-CFG is naturally dependent on the order of the conditioning signals. Intuitively, the first conditional output will have the largest distance from the null output, thus most affecting the final result. In our implementation, our conditioning groups $C$ consist of (1) the empty set (unconditional inference), (2) the clothing-agnostic images $(\{I^0_a, ..., I^{N-1}_a\})$, (3) all clothing-related inputs $(S_g, J_g, M_g)$, and (4) lastly, all remaining conditioning inputs $(\{J^0_p, ..., J^{N-1}_p\}$. We denote the respective weights of each term as $(w_{\emptyset}, w_p, w_g, w_\text{full})$. Empirically, we find this ordering yields the best results.
         
        Overall, we find that controlling sampling via split-CFG not only enhances the frame-wise garment fidelity, but also increases photorealism (FID) the inter-frame consistency of video (FVD), compared to ordinary CFG.  

    \subsection{Progressive Temporal Training}\label{ssec:progressive-training}  
      Our novel progressive temporal training enables up to 64-frame video generation in a single inference run. We first train a base image model from scratch on image data at $512$px resolution and image batches of shape $BxTxHxWxC$, with batch size $B = 8$ and length $T = 1$, for 1M iterations. Then, we inflate the base architecture with temporal blocks and continue training the same spatial layers and new temporal layers with image and video batches with batch size $B = 1$ and length $T = 8$. Video batches are consecutive frames of length $T$ from the same video. After convergence, we double the video length $T$ to 16. This process is repeated until we reach the target length of 64 frames. Each temporal phase is trained for 150K iterations. The benefit of such a progressive process is a faster training speed and better multi-frame consistency. Additional details are provided in the Supplementary.

    \subsection{Joint Image and Video Training}\label{ssec:joint-training} 
    
\begin{figure}
  \centering
  \includegraphics[width=\linewidth]{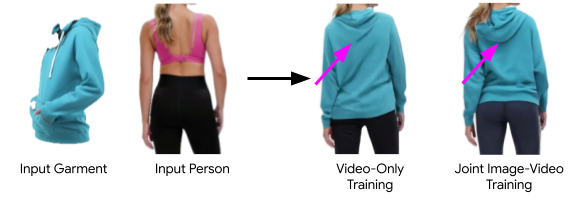}
  \vspace{-20pt}
  \caption{ Joint Training Ablation. Joint image and video training improves the realism of occluded views.}
  \label{fig:joint-training}
  \vspace{-10pt}
\end{figure}

    Training the temporal phases solely with video data, which is much more limited in scale compared to image data, would disregard the image dataset entirely after the pretraining phase. We observe that video-only training in the temporal phases sacrifices image quality and fidelity for temporal smoothness. To combat this issue, we train the temporal phases jointly with 50$\%$ image batches and $50\%$ video batches. We implement joint training via conditional network branching~\cite{stochastic_depth}, i.e. for image batches, we skip updating the temporal blocks in the network. Unlike temporal masking strategies\cite{vdm, imagen-video}, using conditional network branching allows us to include other temporal blocks (Conv-3D, temporal mixing) in addition to temporal attention. Critically, we also train with either image-only or video-only batches, rather than batches of video with appended images~\cite{vdm, imagen-video}. This improves data diversity and training stability by not constraining the possible batches by the number of available video batches. We observe that improved garment fidelity and multi-view realism, especially for synthesized details in occluded garment regions with joint image-video training compared to video-only training (see Figure~\ref{fig:joint-training}).

\section{Experiments}\label{sec:Results}
    In this section, we describe our datasets (Section \ref{ssec:datasets}), evaluation metrics (Section \ref{ssec:metrics}), and results (Section \ref{ssec:results}). We provide training and inference details in the Supplementary.
    
    \subsection{Datasets}
    \label{ssec:datasets}
       Our image dataset is a collection of publicly-crawled online fashion images, containing 17M paired images of people wearing the same garment in different poses. We also collect a video dataset of over 52K publicly-available fashion videos totalling 3.9M frames, which we use for the temporal training phases. During training, the garment image and person frames are randomly sampled from the same video. For evaluation, we collect a separate dataset of 5K videos, containing person videos paired with garment images from a \textit{different} video. Our custom image and video datasets contain a diverse range of skin tones, body shapes, garments, genders, and motions. We also evaluate on the UBC test dataset~\cite{ubc-dataset} of 100 videos. 
       For both test datasets, we randomly pair a garment frame from each video clip with three distinct other video clips to get swapped try-on datasets.

        \subsubsection{Reproducibility} 
        To promote future work in this area and allow fair comparisons with our method, we plan to release a benchmark dataset, including sample paired person videos, garment images, and corresponding preprocessed inputs. We also analyze a version of our model trained and tested exclusively on publicly-available UBC video data~\cite{ubc-dataset} in Section \ref{ssec:ubc-model}.

    \subsection{Metrics}
    \label{ssec:metrics}
        We evaluate our method using FID~\cite{FID}, FVD~\cite{FVD}, and CLIP~\cite{CLIP} scores in Tables~\ref{tab:ablations} and ~\ref{tab:quantitative-comparison}. FID measures the similarity of the distributions of the predicted and ground-truth frames, which gives a measure of the realism of the generated video frames. FVD  measures temporal consistency of video frames. We compute the CLIP image similarity between the segmented garments of the input garment image and predicted frames. In this way, the CLIP score gives us a measure of try-on garment fidelity.

    \subsection{Results}
    \label{ssec:results}
        We showcase qualitative results of our full method in Figure~\ref{fig:qualitative} and provide more qualitative results in the Supplementary. Fashion-VDM is capable of synthesizing smooth, photorealistic try-on videos on a variety of input garment types, patterns, skin tones, genders, and motions.
    
        \subsubsection{UBC-Only Model}\label{ssec:ubc-model}
        
        In order to provide a fair comparison to other methods~\cite{animate-diff, dreampose, stable-video-diffusion}, we train a version of Fashion-VDM using video data only from the UBC dataset. Similar to other methods, we leverage a pretrained image try-on diffusion models and further train using the publicly-available UBC dataset for the video stages. We show the quantitative results in Table~\ref{tab:quantitative-comparison} and provide further details, discussion, and qualitative examples in the Supplementary.


 
%

\begin{table}[t]
\caption{Quantitative Ablation Studies. For each ablated version of our model, we compute FID, FVD, and CLIP scores using both UBC and our test videos with randomly paired garments. Bolded values indicate the best score in each column.}
\vspace{-10pt}
\resizebox{\linewidth}{!}{
  \begin{tabular}{l|ccc|ccc}
    \toprule
    \multicolumn{1}{c}{} & \multicolumn{3}{c}{\bf{UBC Test Dataset}} & \multicolumn{3}{c}{\bf{Our Test Dataset}} \\
     \textit{} & FID $\downarrow$ & FVD $\downarrow$ & CLIP $\uparrow$ & FID $\downarrow$ & FVD $\downarrow$ & CLIP $\uparrow$ \\
    \midrule
    $w/o$ Split-CFG  & 145 & 687 & 0.745 & 78 &  450 & 0.663 \\
    $w/o$ Joint Training  & 106 & 579 & 0.744 & 96 & 565 & 0.651 \\
    $w/o$ Prog. Training & 102 & 631 & 0.736 & 87 & 824 & 0.651 \\
    $w/o$ Temporal Blocks & 94 & 1019 & 0.739 & 95 & 565 & 0.642 \\
    Ours (Full) & \bf{86} & \bf{515} & \bf{0.752} & \bf{71} & \bf{377} & \bf{0.669} \\
  \bottomrule
  \end{tabular}
}
\vspace{-10pt}
\label{tab:ablations}
\end{table}

\begin{figure}
  \centering
  \includegraphics[width=\linewidth]{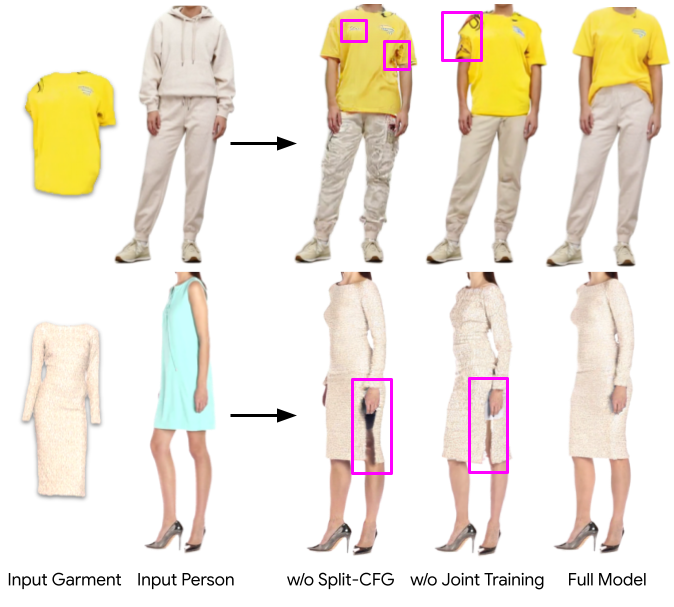}
  \vspace{-20pt}
  \caption{ Garment Fidelity Ablations. We compare our full model with ablated versions without split-CFG and without joint image-video training in terms of garment fidelity. Both split-CFG and joint image-video training improve fine-grain garment details (top row) and novel view generation (bottom row). }
  \vspace{-10pt}
  \label{fig:garment-fidelity}
\end{figure}

\section{Ablation Studies}\label{sec:ablations}

    We ablate each of our design choices with respect to garment fidelity, temporal smoothness, and photorealism.
    We report quantitative results for each ablated version in Table~\ref{tab:ablations}. All components are essential to improving realism (FID), temporal consistency (FVD), and garment fidelity (CLIP). Qualitatively, we find that split-CFG and joint training have the largest effect on person/garment fidelity and overall quality (Figure \ref{fig:garment-fidelity}), while progressive training and temporal blocks affect the temporal smoothness (Figure \ref{fig:temporal-smoothness}). We discuss these effects in detail in the remainder of this section.
    
    \subsection{Split Classifier-Free Guidance}

    Split-CFG improves per-frame person and garment fidelity, thereby improving overall inter-frame temporal consistency and photorealism. In Figure \ref{fig:split-cfg-1}, we compare results generated with different split-CFG weights at inference time. By increasing the person guidance weight $w_p$ from 0 to 1, the realism and identity of the input person are improved. Increasing the full-conditional weight $w_\text{full}$ improves the garment fidelity, but not as much as by increasing the garment weight $w_g$ alone, as in the last column. We provide quantitative split-CFG ablation results in the Supplementary. In the Supplementary, we demonstrate that increasing $w_g$ also increases fine-grain garment details when using a version of our model trained on limited video data. This suggests split-CFG does not require extensive training to be useful and can be impactful in low-resource settings.

    \subsection{Joint Image-Video Training}

    We find that training with video data only in the temporal phases sacrifices garment fidelity compared to the base image model. 
    Training jointly with images and videos increases the fidelity to garment details, even compared to the image baseline, as shown by the improved FID and CLIP scores in Table~\ref{tab:ablations}. The increased access to diverse data with joint image-video training also enables the model to synthesize more plausible occluded regions. For example, as shown in Figure \ref{fig:joint-training}, the jointly trained model is able to generate a hood with more realism than the video-only model.
    

    \subsection{Temporal Blocks}

    As seen in prior works~\cite{vdm, imagen-video}, interleaving 3D-convolution and temporal attention blocks into the 2D UNet greatly improves temporal consistency. Removing temporal blocks entirely causes large temporal inconsistencies. For instance, in the top row of Figure~\ref{fig:temporal-smoothness}, the ablated model without temporal blocks swaps the pants and body shape in each frame.

    \subsection{Progressive Temporal Training}
    \label{ssec:ablation-prog-training}

\begin{figure}
  \centering
  \includegraphics[width=\linewidth]{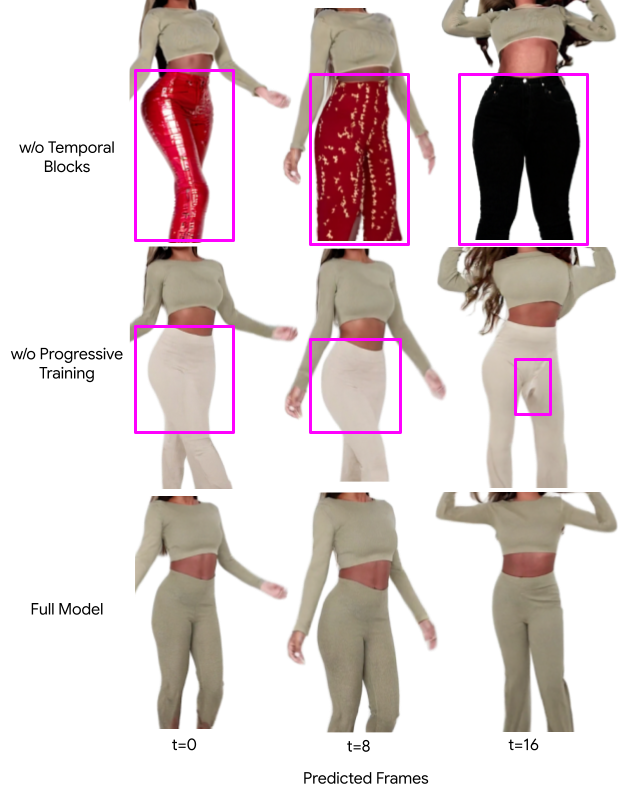}
  \vspace{-20pt}
  \caption{Temporal Smoothness Ablations. We compare video frames generated by our ablated model without temporal blocks (top row) and without progressive training (middle row) to our full model (bottom row). Both ablated versions exhibit large frame-to-frame inconsistencies and artifacts. }
  \label{fig:temporal-smoothness}
  \vspace{-10pt}
\end{figure}

    To ablate our progressive training scheme, we train our image base model directly with 16-frame video batches for the same total number of iterations, but skipping the 8-frame training phase entirely. Progressive training enables more temporally smooth results with the same number of training iterations. This is supported by our quantitative findings in Table~\ref{tab:ablations}, which indicates worse FVD when not doing progressive training. Qualitatively, in Figure~\ref{fig:temporal-smoothness}, the non-progressively trained model in the middle row exhibits temporal artifacts in the pants region and intermittently merges the pant legs into a skirt. We hypothesize that, given limited training iterations, it is easier to learn temporal consistency well across a small number of frames. Then, to transfer that knowledge to larger temporal windows only requires minimal additional training. 
      
\section{Comparisons to State-of-the-Art}\label{sec:results}
    
    We qualitatively and quantitatively compare our method to the state-of-the-art in diffusion-based try-on and animation, as no previous diffusion-based video try-on baselines with publicly-available code currently exist: (1) TryOn Diffusion~\cite{tryondiffusion} (2) MagicAnimate~\cite{magic-animate}, and (3) Animate Anyone~\cite{animate-anyone}. For (1), we generate try-on results in a frame-by-frame manner for each input frame to generate a video. For (2) and (3), we first generate a single try-on image from the first input frame and garment image using TryOn Diffusion, then use the extracted poses from the input frames to animate the result. In addition, we provide user survey results in the Supplementary.

    \subsection{Qualitative Results:} 
    
    We qualitatively compare Fashion-VDM to the baseline methods in Figure~\ref{fig:qualitative-comparisons}. In the top and bottom rows, we show how other methods exhibit large artifacts with large pose changes. In these examples, baseline methods struggle to preserve garment details and hallucinate plausible occluded views. Plus, both MagicAnimate and Animate Anyone create an overall cartoon-like appearance.
    
    In our supplementary video results, we observe that frame-by-frame TryOn Diffusion results exhibit lots of flickering and garment inconsistencies. MagicAnimate fails to preserve the correct background and also does not maintain a consistent garment appearance througout the video. Animate Anyone also exhibits garment temporal inconsistency, especially with large viewpoint changes, and the human motion has an unrealistic, warping effect. Overall, Fashion-VDM synthesizes more natural-looking garment motion, such as folding, wrinkling, and flow, and better preserves garment appearance.
    
    \subsection{Quantitative Results:} 
    
    We compute FID scores on from 300 16-frame videos of the UBC dataset and on 300 16-frame videos of our custom video test dataset. For both datasets, we compute FVD scores and CLIP on 100 distinct 16-frame videos.
    The results are displayed in Table~\ref{tab:quantitative-comparison}. In our experiments, Fashion-VDM surpasses all baselines in both image quality (FID), video quality (FVD), and garment fidelity (CLIP). Although the UBC-only model excels in terms of all UBC metrics, we qualitatively observe over-smoothing and worse garment detail preservation, compared to the full version trained on our larger, more diverse video dataset.


\begin{table}[t]
\caption{Quantitative Comparisons. We compare Fashion-VDM to the baseline methods using the UBC test dataset~\cite{ubc-dataset} and our test dataset of internet videos. Fashion-VDM quantitatively outperforms other methods on all metrics.}
\vspace{-10pt}
\resizebox{\linewidth}{!}{
  \begin{tabular}{l|ccc|ccc}
    \toprule
    \multicolumn{1}{c}{} & \multicolumn{3}{c}{\textbf{UBC Test Dataset}} & \multicolumn{3}{c}{\textbf{Our Test Dataset}} \\
     \textit{} & FID $\downarrow$ & FVD $\downarrow$ & CLIP $\uparrow$ & FID $\downarrow$ & FVD $\downarrow$ & CLIP $\uparrow$ \\
    \midrule
    TryOn Diffusion & 94 & 1019 & 0.739 & 95 & 960 & 0.663 \\
    Magic Animate & 155 & 1861 & 0.702 & 97 & 694 & 0.642 \\
    Animate Anyone & 118 & 819 & 0.727 & 112 & 468 & 0.629 \\
    Ours (Full) & 86 & 515 & 0.752 & \bf{71} & \bf{377} & \bf{0.669} \\
    \midrule
    \midrule
    Ours (UBC-Only) & \bf{39} & \bf{172} & \bf{0.749}& 129 & 949 & 0.657 \\
    \bottomrule
  \end{tabular}
}
\vspace{-15pt}
\label{tab:quantitative-comparison}
\end{table}





\section{Limitations and Future Work}
The main limitations of Fashion-VDM include inaccurate body shape, artifacts, and incorrect details in occluded garment regions. See examples and further discussion in the Supplementary. Improbable details may be hallucinated in unseen garment regions, because the input image only shows one view of the garment. Future work might consider multi-view conditioning and individual person customization for improved garment and person fidelity. Other errors include minor aliasing for fine-grained patterns. Finally, our method does not simulate exact physical cloth dynamics, but rather realistic video try-on visualization. Establishing physics could be a great next step.

\section{Discussion}
We present Fashion-VDM, a diffusion-based video try-on model. Given an input garment image and person video, Fashion-VDM synthesizes a try-on video with the input garment fitted to the person in motion, maintaining realistic details and fabric dynamics. We show qualitatively and quantitatively that our method significantly surpasses existing state-of-the-art diffusion-based image try-on and animation methods. 

\section{Ethics Statement}
    While we believe our research creates a positive contribution to the research community by advancing the state-of-the-art in generative video diffusion, we also condemn its potential for misuse, including any spreading misinformation or manipulating human content for malicious purposes. While our method is trained on public data containing identifiable humans, we will not release any images or videos containing personally identifiable features, such as faces, tattoos, or logos to protect the privacy of these individuals.

\bibliographystyle{ACM-Reference-Format}
\bibliography{main}

\appendix
\begin{figure*}
  \centering
  \includegraphics[width=\linewidth]{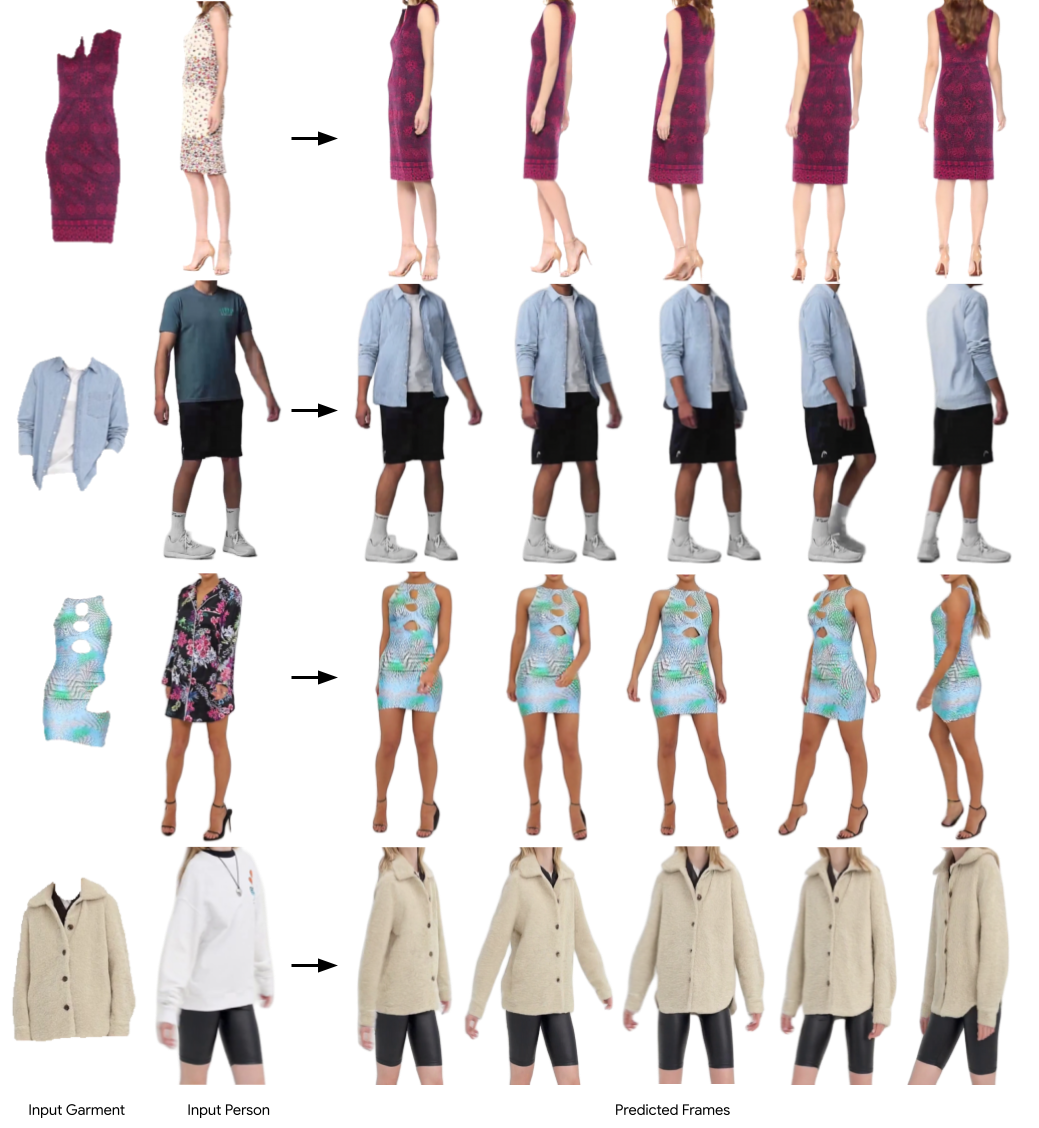}
  \vspace{-10pt}
  \caption{Qualitative Results. We showcase video try-on results generated by Fashion-VDM using randomly paired person-garment test videos from the UBC dataset~\cite{ubc-dataset} and our own collected test dataset. Note that the input garment image and input person frames come from different videos.}
  \label{fig:qualitative}
\end{figure*}
\begin{figure*}
  \centering
  \includegraphics[width=0.9\linewidth]{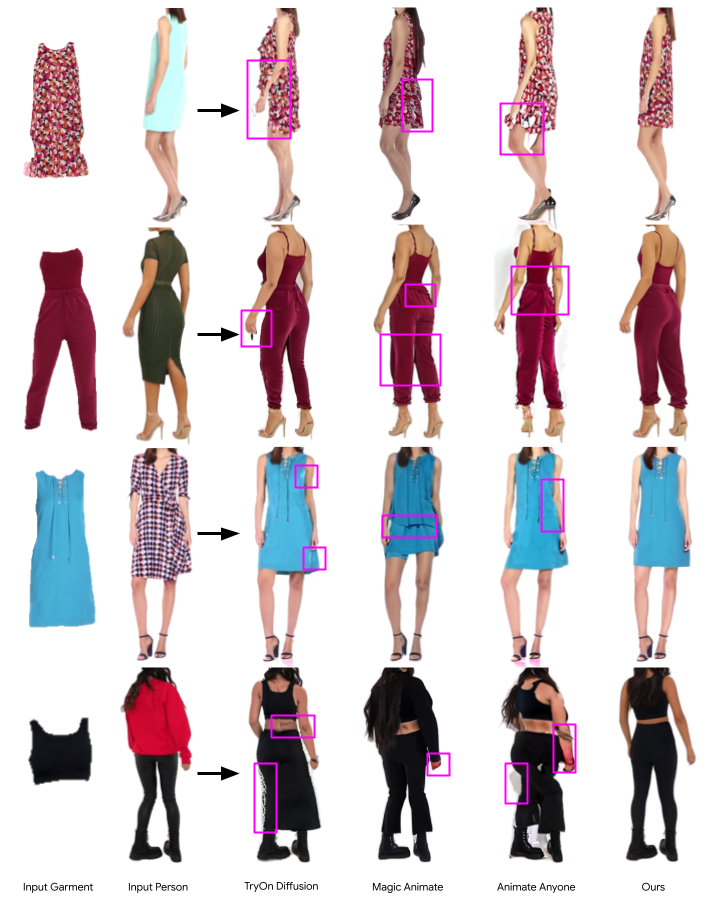}
  \vspace{-10pt}
  \caption{Qualitative Comparisons. Fashion-VDM outperforms past methods in garment fidelity and realism. Especially in cases of large disocclusion, our method synthesizes more realistic novel views.}
  \label{fig:qualitative-comparisons}
\end{figure*}
\clearpage

\setcounter{section}{0}
\begin{center}
    \textbf{\Large Supplementary Material}
\end{center}

\section{Progressive Training Details}
\label{ssec:progressive-training-details}
    \begin{figure*}
  \centering
  \includegraphics[width=\linewidth]{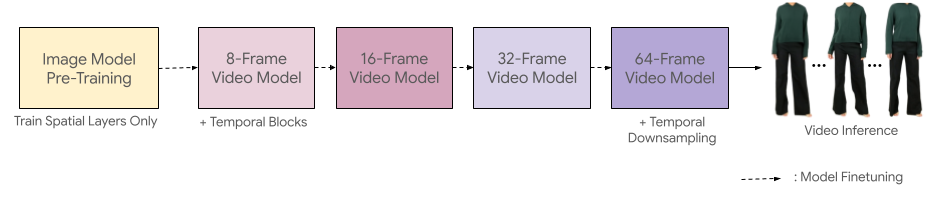}
  \vspace{-10pt}
  \caption{ Progressive Training Strategy. Fashion-VDM is trained in multiple phases of increasing frame length. We first pretrain an image model, by training only the spatial layers on our image dataset. In subsequent phases, we train temporal and spatial layers on increasingly long batches of consecutive frames from our video dataset.}
\label{fig:progressive-training}
\end{figure*}

     The overall progressive temporal training strategy is depicted in Figure~\ref{fig:progressive-training}. We first train a base image model from scratch on image data at $512$px resolution and batch size 8 for 1M iterations. Then, we inflate the base architecture with temporal blocks and continue training the model using our joint image-video training strategy. In these temporal training phases, half of the batches are from the image dataset and the other half are batches of consecutive video frames from the video dataset. When training with an image batch, we skip the temporal blocks entirely in the forward and backward passes.  At each successive phase of temporal training, we initialize the model from the previous phase's checkpoint and double the training video length: $8 \rightarrow 16 \rightarrow 32 \rightarrow 64$. We train each temporal training phase for ~150K iterations. Once the video length becomes prohibitively large in memory at 64-frames, we introduce temporal downsampling and upsampling layers to the model. At test time, our model generates $512 \times 384$px videos up to 64-frames in one inference pass with a single network.

\subsection{Training and Inference Details} \label{ssec:implementation-details} 
    
    We train our model on 16 TPU-v4's for approximately 2 weeks, including all training phases. Our image baseline model is trained for 1M iterations with a batch size of 8 and resolution $512 \times 384$px using the Adam optimizer with a linearly decaying learning rate of $1e^{-4}$ to $1e{-5}$ over 1M steps and 10K warm-up steps. Each phase of progressive temporal training is initialized from the previous checkpoint and trained for 150K iterations, following the order of phases described in the Section ~\ref{ssec:progressive-training-details}. For all phases, we incorporate dropout for each conditional input independently 10\% of the time. We train with an L2 loss on $\epsilon$.

    During inference, we use the DDPM sampler \cite{ddpm} with 1000 refinement steps. Each video takes approximately 8 minutes to synthesize with split-CFG and 5 minutes without split-CFG. 
    
    \begin{figure}
  \centering
  \includegraphics[width=\linewidth, trim={0 1cm 0 0}]{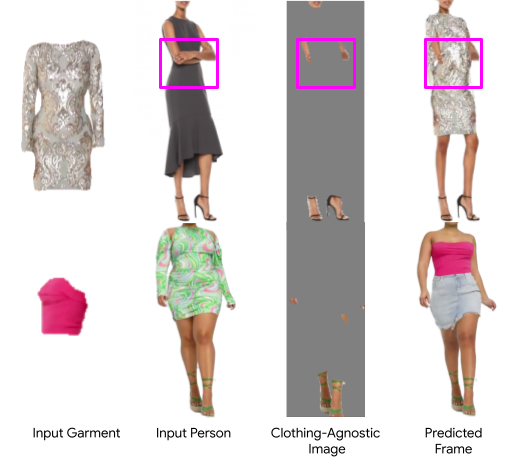}
  \caption{ \textbf{Failure Cases.} Errors in the person segmentation may lead to artifacts (top row). Fashion-VDM may incorrectly represent body shape (bottom row).}
  \label{fig:failure-cases}
\end{figure}

\section{Examples of Failure Cases}
    We show two examples of failure cases of our method in Figure~\ref{fig:failure-cases}. In row 1, we show artifacts that appear in the body/garment boundary, due to an imperfect person segmentation in the clothing-agnostic image. Imperfect segmentation is a common cause of such artifacts, and may also incorrectly leak regions from the original garment. In our human evaluation (Section~\ref{ssec:user-study}), 10/17 videos that were failed had agnostic errors.  In general, although our preprocessing methods are state-of-the-art, other types of preprocessing errors occur limit the quality of Fashion-VDM. In total, 70\% of videos not chosen by human raters had errors in one or more inputs. As shown in row 2, body shape misrepresentation (e.g. slimming) occurs, because the clothing-agnostic images remove all body parts, besides hands, feet, and head, thus they do not include detailed information about body size.

\section{Split-CFG Weights Ablations}
    We quantitatively evaluate our choice of split-CFG weights for both datasets on a held-out validation set. The results are shown in Table \ref{tab:split-cfg-ablations}. Calibrating these weights correctly is not only beneficial to preserving garment fidelity, as shown by the FID score, but also increasing temporal consistency, as shown by the FVD score. Intuitively, by increasing the similarity of the output garment to the input garment, there is less allowed variability in the appearance of each frame, thus increased temporal smoothness. Based on these results, we employ weights $(1, 1, 3, 1)$ for UBC and weights $(1, 1, 1, 1)$ for our test dataset.
    
    \begin{table}[t]
\resizebox{\linewidth}{!}{
  \begin{tabular}{l|ccc|ccc}
    \toprule
    \multicolumn{1}{c}{} & \multicolumn{3}{c}{\textbf{UBC Test Dataset}} & \multicolumn{3}{c}{\textbf{Our Test Dataset}} \\
     \textit{$(w_{\emptyset}, w_p, w_g, w_c)$} & FID $\downarrow$ & FVD $\downarrow$ & CLIP $\uparrow$ & FID $\downarrow$ & FVD $\downarrow$ & CLIP $\uparrow$ \\
    \midrule
    (1, 0, 1, 1)  & 136 & 1053 & 0.712 & 126 & 779 & 0.632 \\
    (1, 1, 1, 1)  & 95 & 644 & 0.748 & $\bf{52}$ & $\bf{242}$ & 0.667 \\
    (1, 1, 1, 3)  & 100 & 653 & 0.745 & 84 & 588 & 0.664 \\
    (1, 1, 3, 3)  & 99 & 454 & 0.756 & 55 & 262 & 0.662 \\
    (1, 3, 3, 3) & 104 & 481 & 0.763 & 63 & 284 & $\bf{0.671}$ \\
    (1, 1, 3, 1)  & $\bf{62}$ & $\bf{253}$ & $\bf{0.770}$ & 76 & 385 & 0.661 \\
  \bottomrule
  \end{tabular}
}
\caption{Quantitative Ablation of Split-CFG Weights. We compute FID, FVD, and CLIP scores of our full model using different split-CFG weights.}
\label{tab:split-cfg-ablations}
\end{table}

\subsection{User Study}
\label{ssec:user-study}
    \begin{table*}[t]
\resizebox{\linewidth}{!}{
  \begin{tabular}{l|ccc|ccc|ccc}
    \toprule
    \multicolumn{1}{c}{} & \multicolumn{3}{c}{\textbf{UBC Test Dataset}} & \multicolumn{3}{c}{\textbf{Our Test Dataset}} \\
      & Video Smoothness $\uparrow$ & Person Fidelity $\uparrow$& Garment Fidelity $\uparrow$ & Video Smoothness $\uparrow$& Person Fidelity $\uparrow$ & Garment Fidelity $\uparrow$ \\
    \midrule
    TryOn Diffusion & 0.01 & 0.00 & 0.00 & 0.03 & 0.01 & 0.00 \\
    Magic Animate & 0.03 & 0.00 & 0.03 & 0.02 & 0.01 & 0.00 \\
    Animate Anyone & 0.03 & 0.03 & 0.03 & 0.04 & 0.01 & 0.05 \\
    Ours (Full) & 0.93 & 0.97 & 0.94 & 0.91 & 0.96 & 0.95 \\
  \bottomrule
  \end{tabular}
}
\caption{User Study. Our study indicates that users overwhelmingly prefer Fashion-VDM to other baselines in terms of video smoothness, person fidelity, and garment fidelity on both test datasets. }
\label{tab:user-study}
\end{table*}

    In addition to qualitative and quantitative evaluations, we perform user studies for our state-of-the-art comparisons. The results are shown in Table~\ref{tab:user-study}. Our user studies are conducted by 5 human raters who are unfamiliar with the method. For each sample, the raters were asked to select which video performs best in each category: temporal smoothness, garment fidelity to the input garment image, and person fidelity to the input person video. The scores on both UBC test dataset and our test dataset reported are fraction of total votes divided by the total number of videos. Fashion-VDM exceeds other methods on all three user preference categories for both datasets.

\subsection{UBC-Only Model}
   \begin{figure}
      \centering
      \includegraphics[width=\linewidth]{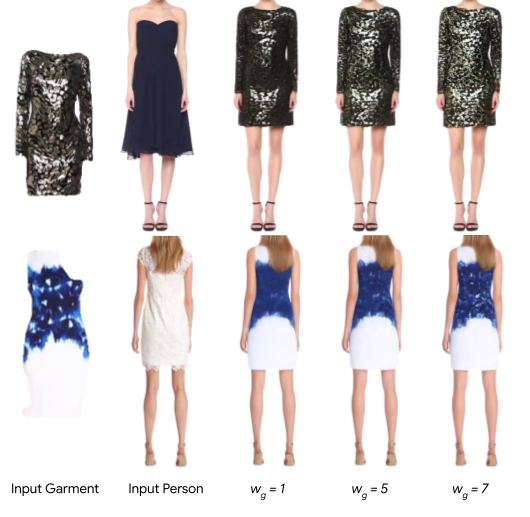}
      \vspace{-10pt}
      \vspace{-10pt}
      \caption{ \textbf{Split-CFG Ablation with UBC-Only Model.} When Fashion-VDM is trained on the limited UBC dataset only, we observe overfitting to the largely plain garments in the UBC train dataset. However, we find that increasing garment image guidance ($w_g$) in split-CFG significantly increases garment details.}
    \label{fig:ubc-only-split-cfg}
    \end{figure}
    
   \begin{figure*}
    \centering
      \includegraphics[width=\linewidth]{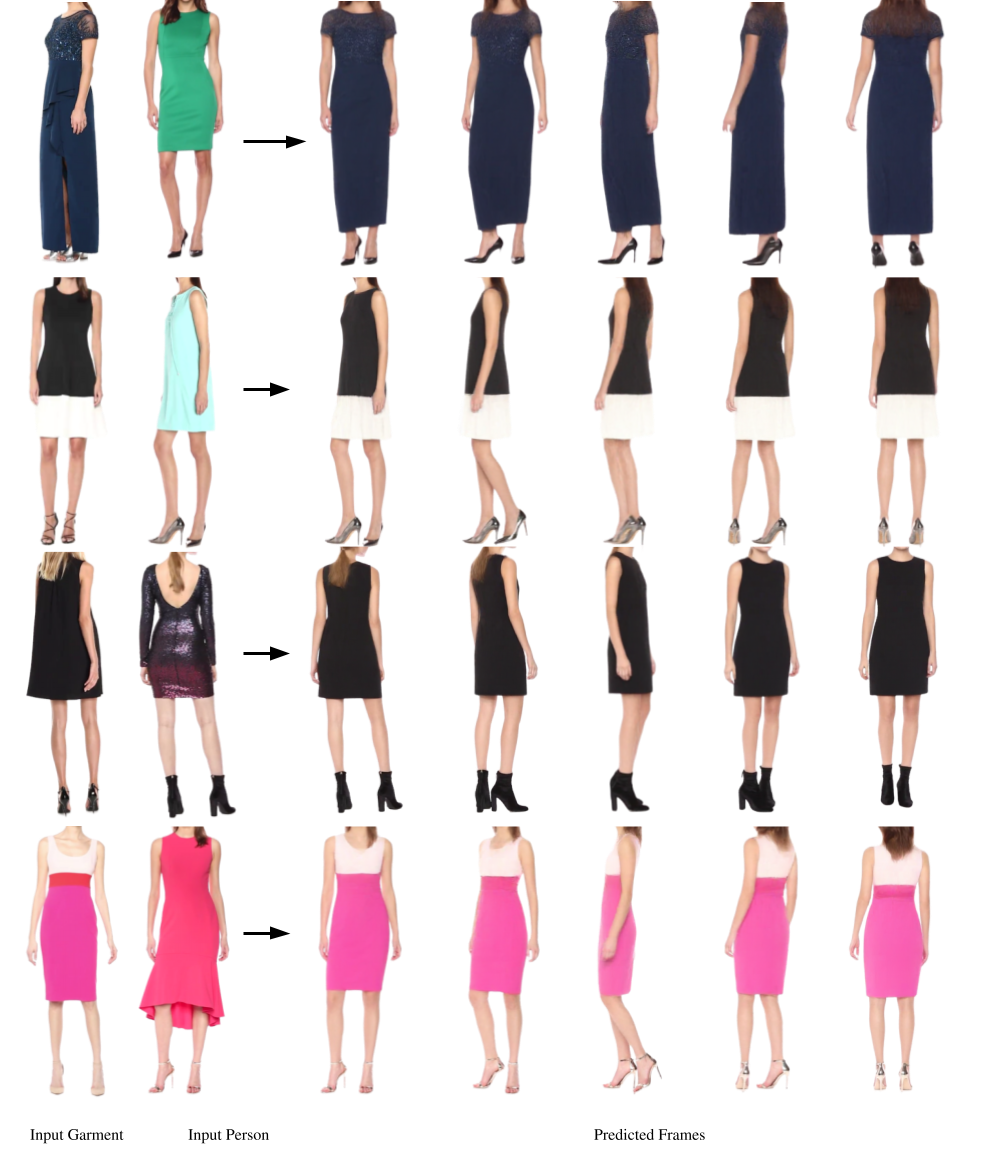}
      \caption{ \textbf{Qualitative Results for UBC-Only Model.} Our model trained only on UBC data generates temporally consistent, smooth try-on videos for plain and simple patterned garments, but struggles to preserve intricate patterns and complex garment shapes.}
      \label{fig:qualitative-ubc-only}
    \end{figure*} 
   We initialize this model from our pretrained image model, which is comparable to an open source image diffusion model, like Stable Diffusion~\cite{stable-diffusion}, which are trained on even larger image datasets, including LAION 5B~\cite{laion5b}. We then train progressively using both image data and UBC video data, following the same progressive training scheme as the full model.
    
   The UBC-only model exceeds all baselines on the UBC test dataset quantitatively, but is qualitatively worse at preserving intricate garment details and patterns. This is expected, given the limited size and lack of diversity of UBC training dataset. However, we discovered that increasing the split-CFG garment weight significantly improves lost garment details, even more so than with the full model. We qualitatively show this in Figure~\ref{fig:qualitative-ubc-only}. This implies that when training with limited data, split-CFG becomes even more crucial to preserving the conditioning image details.

    We provide qualitative examples generated by our model trained only on the UBC dataset~\cite{ubc-dataset} in Figure~\ref{fig:qualitative-ubc-only}. While the results are still smooth and temporally consistent, the model struggles to maintain complex patterns and garment shape details. This is likely due to overfitting to the limited size and scope of the UBC training dataset, consisting of 500 videos of women in dresses.

   \begin{figure*}
      \centering
      \includegraphics[width=0.9\linewidth, trim={0 0 1cm 0}]{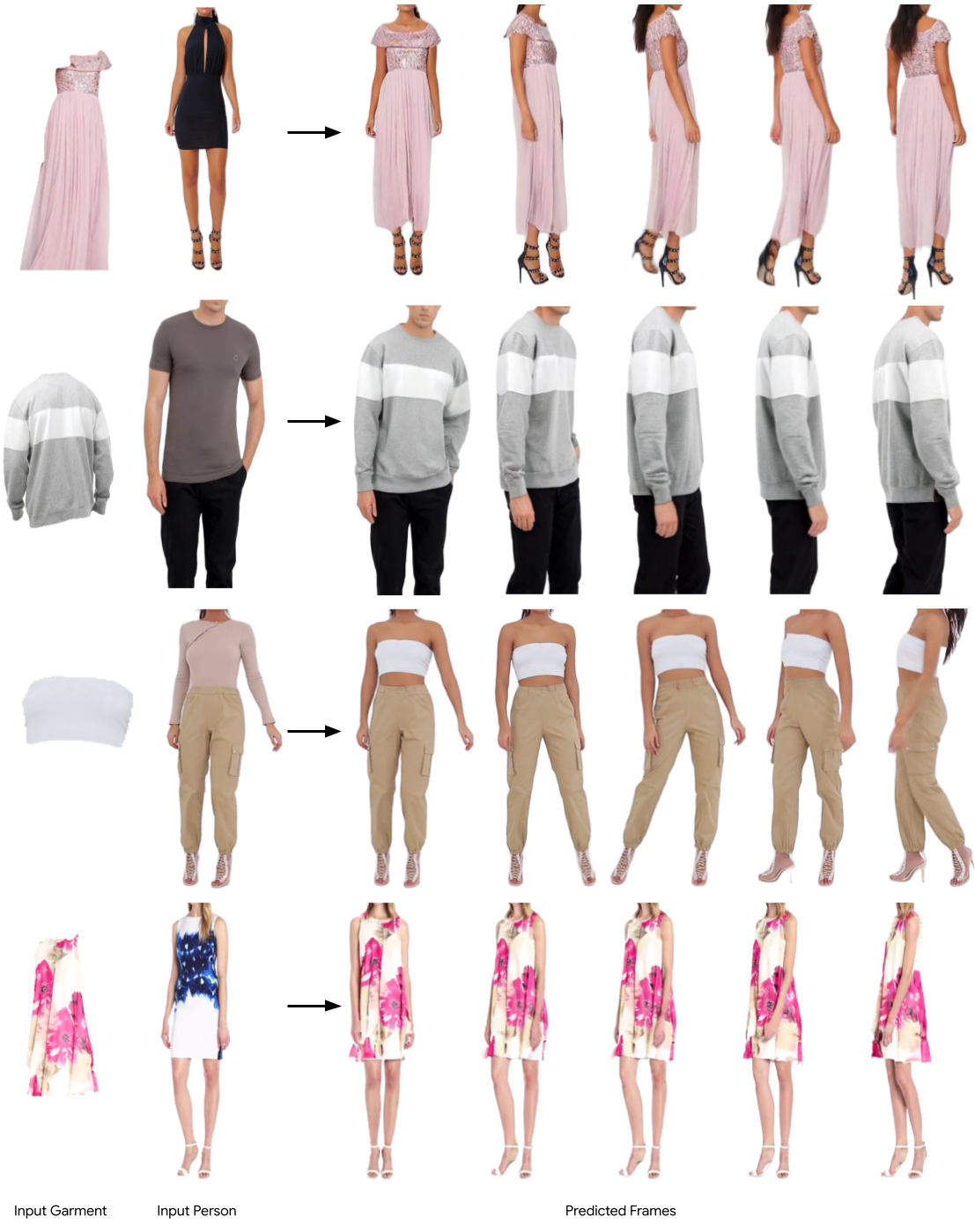}
      \caption{ \textbf{Additional Qualitative Results.} We showcase video try-on results generated by Fashion-VDM using swapped test videos from the UBC dataset~\cite{ubc-dataset} and our own collected test dataset. Note that the input garment image and input person frames come from different videos.}
      \label{fig:qualitative-supp}
    \end{figure*}

\end{document}